\documentclass[runningheads]{llncs}
\usepackage{graphicx}
\usepackage{float}

\begin{document}
\title{xAI-CycleGAN, a Cycle-Consistent Generative Assistive Network}
\author{Tibor Sloboda\orcidID{0000-0001-6817-6297}\and\\
Lukáš Hudec\orcidID{0000-0002-1659-0362}\and\\
Wanda Benešová\orcidID{0000-0001-6929-9694}}

\institute{Vision \& Graphics Group, FIIT STU\\
\email{xslobodat2@stuba.sk}\\
\url{https://vgg.fiit.stuba.sk/}}

\maketitle              
\begin{abstract}
In the domain of unsupervised image-to-image transformation using generative transformative models, CycleGAN~\cite{Zhu2017UnpairedNetworks} has become the architecture of choice. One of the primary downsides of this architecture is its relatively slow rate of convergence. In this work, we use discriminator-driven explainability to speed up the convergence rate of the generative model by using saliency maps from the discriminator that mask the gradients of the generator during backpropagation, based on the work of Nagisetty et al.~\cite{Nagisetty2020XAI-GAN:Systems}, and also introducing the saliency map on input, added onto a Gaussian noise mask, by using an interpretable latent variable based on Wang M.'s Mask CycleGAN~\cite{Wang2022MaskVariable}. This allows for an explainability fusion in both directions, and utilizing the noise-added saliency map on input as evidence-based counterfactual filtering~\cite{Vermeire2022ExplainableCounterfactual}. This new architecture has much higher rate of convergence than a baseline CycleGAN architecture while preserving the image quality.
\end{abstract}

\section{Introduction}
Unsupervised image-to-image transformation using generative transformative models has gained significant attention in recent years. Among the various architectures developed, CycleGAN has emerged as a popular choice due to its ability to learn transformations between unpaired datasets. However, one of the major drawbacks of CycleGAN is its slow convergence rate. In this work, we propose a method to combine the approaches of two authors with unrelated contributions to create a novel approach and architecture that significantly speeds up converge of CycleGAN.

CycleGAN is an autoencoder-like architecture consisting of two pairs of generators and discriminators. The generators aim to generate realistic images, while the discriminators try to distinguish between real and fake images. The cyclic nature of the network allows for unsupervised learning by passing the generated images through the other generator to ensure consistency with the original input.

Building upon CycleGAN, we incorporate the concepts from two existing approaches: Mask CycleGAN and discriminator-driven explainability. Mask CycleGAN introduces an interpretable latent variable using hard masks on the input~\cite{Wang2022MaskVariable}, allowing for selective image transformation. However, it warns against using soft masks due to potential information leakage, which we actually take advantage of in our work.

We combine this approach with discriminator-driven explainability, inspired by Nagisetty et al.~\cite{Nagisetty2020XAI-GAN:Systems} This approach involves using saliency maps from the discriminator to mask the gradients of the generator during backpropagation, which we also use in the latent interpretable variable as a mask on input.

\paragraph*{Our Contribution}
In our proposed method, we combine the interpretable latent variable from Mask CycleGAN with the explainability-driven training of the generator. We use a soft mask on the input, which includes Gaussian noise, and apply the saliency maps from the discriminator as a mask on the gradients during backpropagation as well as added to the mask on the input. Additionally, we adjust the saliency maps based on the performance of the discriminator, allowing for adaptive filtering of information.

Our architecture, named xAI-CycleGAN, demonstrates a significantly higher convergence rate compared to the baseline CycleGAN while preserving image quality. We evaluate the performance using the horses and zebras dataset and qualitatively compare the results. The initial experiments show promising results, with xAI-CycleGAN producing high-quality images even at early epochs of training.

\section{Related work}
\subsection{CycleGAN}
Our work is based on CycleGAN, the autoencoder-like architecture that contains two pairs of generator and discriminator, where the generator and discriminator compete in a zero-sum game~\cite{Zhu2017UnpairedNetworks}. The generator attempts to improve by producing continuously more convincing images, while the discriminator attempts to improve by discerning which from a pair of two images is fake, and which one is real.

It is based on classical GAN models~\cite{goodfellow2020generative} which have shown impressive results in both generative and image editing tasks\cite{denton2015deep,zhu2016generative,Zhu2017UnpairedNetworks}.

The cyclic nature of the network that makes it possible to use it in unsupervised, unpaired scenarios comes in with the cycle pass of the fake image through the other generator. Given an image of domain $X$ and the generator $X \rightarrow Y$, we produce a fake image $Y'$ and then use the $Y \rightarrow X$ generator to translate $Y'$ back to $X'$, the cycled image, which needs to be consistent with the original image $X$.

\subsection{Mask CycleGAN}
Mask CycleGAN introduces an interpretable latent variable using hard masks on the input and fusing them together, using a $1 x 1$ 2D convolutional layer, and then adds the inverse of the mask to the output, which produces our resulting image.

The method introduces new hyperparameters that balance the importance of the mask on input as well as the effect of a new mask discriminator's loss on the total loss. This allows the author to isolate which parts of the image are to be converted, while the rest remains untouched on the basis of the hard masks. The mask discriminator is introduced to prevent the boundaries of the hard mask to be used as evidence of the image being fake.

The author warns against the use of soft masks as they may cause features to leak into the encoder, and they have therefore avoided those.

\subsection{Discriminator-driven explainability assisted training}
Nagisetty et al. proposes a new approach where the discriminator provides a richer information to the generator for it to learn faster and need less data to achieve similar results~\cite{Nagisetty2020XAI-GAN:Systems}. The authors achieved this by generative explanation matrices based on the generated image, the discriminator’s prediction, and the discriminator model itself.

The explanation matrix is used to mask the gradient of the generator output with respect to the loss $\nabla_{G(z)}$, before the gradient is backpropagated, hence changing how the generator learns, and filtering out unimportant information:

\begin{equation}
    \nabla^\prime _{G(z)} = \nabla_{G(z)} + \alpha \nabla_{G(z)} \odot M
\end{equation}

\noindent
Here $\nabla_{G(z)}$ represents the generator gradient based on the discriminator classification, $M$ is the explanation matrix and $\alpha$ is a hyperparameter with a small value to dampen the magnitudes of the gradient before a Hadamard product is computed between the gradients and the explanation matrix, in order to modify the resulting gradient only by small increments.

This method was originally only designed for classical GANs~\cite{goodfellow2020generative} but can be expanded to any such architecture based off of them.

\section{Proposed Method}
Beginning with a base CycleGAN, we first introduce the interpretable latent variable using a soft mask to distill semantically significant features into the encoder layer for the mask. Wang M.~\cite{Wang2022MaskVariable} in the original paper Mask CycleGAN warns that a soft mask may cause unwanted information leaking from the gradient during backpropagation into the mask encoder.

\subsection{Explanation map application}
While this was considered a downside in the original paper, we exploit the soft mask leakage feature by introducing Lambda-adaptive explainability assisted training based on Nagisetty et al.~\cite{Nagisetty2020XAI-GAN:Systems} which masks the gradients during backpropagation with explanation or saliency maps from the discriminator. We combine this with a soft mask on input that contains Gaussian noise, with a mean equal to 1.0 and with a standard deviation of 0.02, to excessively express this denoted downside of the approach and distill these explanations or saliency maps through the learned parameters of the network into the mask encoder, but otherwise keep the mask mostly transparent to the network, and remove the need for a mask during inference.

For our explainability approach to produce explanation maps from the discriminator, we use saliency as it has shown to be the most effective based on the rate of convergence. The explanation maps are adjusted by a new pair of hyperparameters, $\lambda_{a}$ and $\lambda_{b}$ which are based on the performance of the discriminator:

\begin{equation} \label{eq:lambdafunc}
    \lambda_{a, b} = min(1, 4(min(0.5, x) - 0.5)^\gamma)
\end{equation}

\noindent
The $x$ represents the loss of the discriminator, and $\gamma$ adjusts the slope of the function. The function has a maximum value of $1.0$ when the loss of the discriminator equals to $0.0$, or a minimal value of $0.0$ when the loss of the discriminator is equal to $0.5$, because that in our case represents a discriminator that cannot distinguish between a fake and real image, so it cannot be trusted.

We use explanation maps from the primary discriminator to mask the gradients and explanation maps from the mask-discriminator, identical to the one in Mask CycleGAN, to add to the input mask.

In the cases where the explanation map becomes significantly dampened by $\lambda_b$, due to the mask on input being Gaussian noise, it becomes mostly indistinguishable from the noise itself and therefore no longer has any impact on the generator, which also additionally serves as a means of augmenting the data by adding random noise. 

\subsection{Generative assistive network}
Just like in the original paper, the mask is concatenated with our image, before being reduced back to 3 channels like the input image using a $1 \times 1$ Conv2D layer. In addition, however, we run the generative process in two passes, where weights are only being updated in the second pass, due to us adding the explanation mask from the mask discriminator network to the mask on input, produced by the first pass, adjusted by a new hyperparameter $\lambda_b$.

The idea behind this approach is that the explanation map from the mask discriminator contains insight on which part of the image contributes to the correct identification of the true class of the generated image, and which parts are detrimental. Normally, this information is backpropagated through the network, and weights are adjusted based on the error. Here, we take a two-sided approach of counterfactual-suppression or prefactual-suppression using the mask to either reduce the initial errors made by the generator, or to exaggerate them for the discriminator to then provide better guidance to the generator.

Both approaches need to be tested for feasibility, but in either case, this re-defines the idea of a generative adversarial network to a generative assistive network, as the explanations from the discriminator are helping the generator learn, and promote the generator into tricking the discriminator, rather than the typical competition analogy.

The idea is supported by an already existing well-developed method of counterfactual suppression in image classification networks, where removing, masking or hiding parts of the image helps correct the predicted image class~\cite{Vermeire2022ExplainableCounterfactual}. This, however, can also be used to produce adversarial examples, which would in turn cause a misclassification of an otherwise correctly classified image. It is possible that the generator may learn to use this advantage to generate examples where the discriminator will always be tricked into thinking the generator is producing good images.

\subsection{Architecture technical parameters}
The initial setup of the neural network before adjustments were made based on the results contains several hyperparameters. The model was implemented in PyTorch and based on the original CycleGAN implementation with added skip-connections and modified discriminators and activation functions.

The generator network consists of two initial mask encoder layers for the mask and inverted mask to be combined with the image, followed by an initial feature extraction convolution layer that maintains the shape of the encoded image, but results in \verb|ngf| filters, where \verb|ngf| is the hyperparameter that represents the initial filter count before downconvolution.

\begin{figure}[H] \centering

    \includegraphics[width=0.95\textwidth]{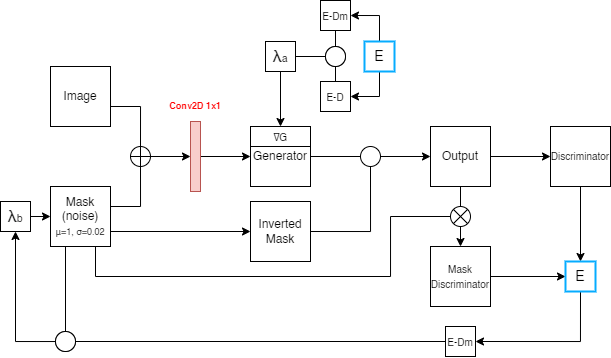}

    \caption{The custom xAI enhanced CycleGAN architecture --- xAI-CycleGAN; The 'E' blocks represent our produced explanation maps that are used to mask the generator gradients after being dampened by $\lambda_a$ and also added to the input, dampened by $\lambda_b$.}

    \label{fig:xaicyclegan}

\end{figure}

Following are three downconvolution steps that half the size using a kernel size 3 with stride 2 and padding 1, while doubling the amount of filters.

Next, we have our latent feature maps, which are transformed with residual blocks that maintain the feature map dimensions. The number of residual blocks can be adjusted using the \verb|num_resnet| hyperparameters.

Finally, we have three upconvolution steps which use transposed convolution layers to achieve dimensions equal to the downconvolution steps at a given level, which allows us to add the downconvolution step to the upconvolution as a skip connection. At the end, we have a convolution layer that reduces the number of filters from \verb|ngf| to 3 as a standard 3-channel image, which can then be visualized.

The discriminators use an alternating set of downconvolution layers with kernel size 4 and stride 2 for halving the dimensions with batch norm disabled, and kernel size 7 while preserving dimensions for feature extraction at various levels of downconvolution. The initial filter size is determined by the hyperparameter \verb|ndf| and doubles with each halving of the spatial dimensions, where the error is the evaluated based on mean square error.

\section{Results}
We used the classical horses and zebras dataset to evaluate the new architecture and approach qualitatively, by comparing the quality of the images at the same training step between baseline CycleGAN and our improved architecture.

\begin{figure}[H] \centering

    \includegraphics[width=0.5\textwidth]{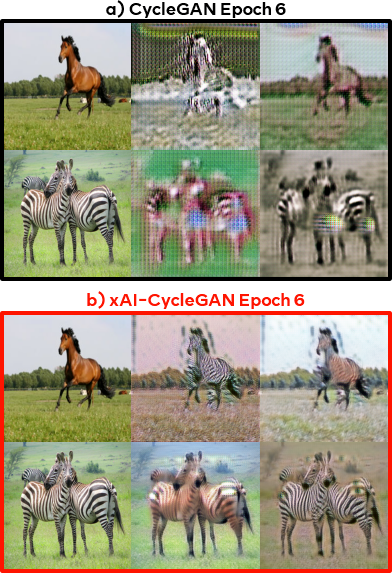}

    \caption{a) The results of ordinary unmodified CycleGAN with the same hyperparameters at epoch=6 as b) xAI-CycleGAN results with added mask and explanation assisted training, and other minor modifications at epoch=6}

    \label{fig:cycvsxaicyc}

\end{figure}

The initial results showed great promise, where translations from \textbf{domain A (horse)} to \textbf{domain B (zebra)} produced excellent quality images even as early as the 6th epoch of training. This highly exceeds the capabilities of an unmodified original CycleGAN.

The loss graphs also reveal that xAI-CycleGAN is able to learn much faster and better initially as opposed to the classic CycleGAN which further supports the qualitative assessment:

\begin{figure}[H] \centering

    \includegraphics[width=0.95\textwidth]{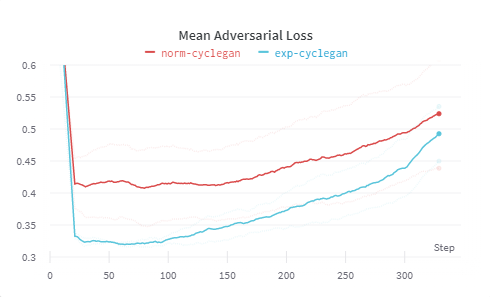}

    \caption{Comparison of mean adversarial losses of an ordinary CycleGAN architecture (norm-cyclegan) against xAI-CycleGAN (exp-cyclegan).}

    \label{fig:cycexpcompare}

\end{figure}

There are still some issues with the approach, and the generator managed to sometimes produce counterfactual examples. If we investigate the results later in the training, we start to see that the generator has learned to take advantage of the explanation maps in order to produce adversarial examples which trick the discriminator:

\begin{figure}[H] \centering

    \includegraphics[width=0.5\textwidth]{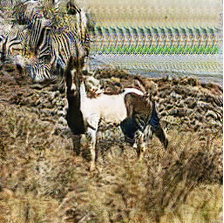}

    \caption{Sample of a produced image where the generator created a patch of zebra-like texture to trick the discriminator.}

    \label{fig:advexptrick}

\end{figure}

This is likely due to the initial choice of explanation maps, which are currently using the saliency map approach, where the activations from the output layer of the discriminator, based on the loss, backpropagate to the input where we are able to produce a saliency map of importance values to pixels.

\section{Discussion}
This approach introduces a new possibility and method of training an unsupervised transformative generative model, in this case CycleGAN, significantly boosting the convergence rate and the speed at which it produces good quality transformations, albeit at the cost of some issues that still need to be addressed such as the presence and possibility of the generator producing counterfactuals later during training.

We believe that with additional specifically-tailored loss function, it is possible to entirely eliminate these counterfactual examples and make the approach completely reliable.

With concerns over the energy use of training networks, especially with a large amount of data and parameters, this is a positive step forward to increasing the training efficiency in one domain of deep learning.

\newpage
\bibliographystyle{splncs04}
\bibliography{references}

\end{document}